\pdfoutput=1
\documentclass[10pt,twocolumn,letterpaper]{article}

\usepackage{cvpr}
\usepackage{times}
\usepackage{epsfig}
\usepackage{graphicx}
\usepackage{amsmath}
\usepackage{amssymb}
\newcommand{\ignore}[1]{}
\newcommand\blfootnote[1]{%
\begingroup
\renewcommand\thefootnote{}\footnote{#1}%
\addtocounter{footnote}{-1}%
\endgroup
}

\usepackage[pagebackref=true,breaklinks=true,letterpaper=true,colorlinks,bookmarks=false]{hyperref}

\cvprfinalcopy 

\def\cvprPaperID{xxxx} 

\begin{document}

\title{Defending Against Model Stealing Attacks with Adaptive Misinformation}

\author{Sanjay Kariyappa\\
Georgia Institute of Technology\\
Atlanta GA\\
{\tt\small sanjaykariyappa@gatech.edu}
\and
Moinuddin K Qureshi\\
Georgia Institute of Technology\\
Atlanta GA\\
{\tt\small moin@gatech.edu}
}

\maketitle

\begin{abstract}
   Deep Neural Networks (DNNs) are susceptible to model stealing attacks, which allows a \emph{data-limited} adversary with no knowledge of the training dataset to clone the functionality of a target model, just by using black-box query access. Such attacks are typically carried out by querying the target model using inputs that are synthetically generated or sampled from a surrogate dataset to construct a labeled dataset. The adversary can use this labeled dataset to train a clone model, which achieves a classification accuracy comparable to that of the target model. We propose ``Adaptive Misinformation" to defend against such model stealing attacks. We identify that all existing model stealing attacks invariably query the target model with Out-Of-Distribution (OOD) inputs. By selectively sending incorrect predictions for OOD queries, our defense substantially degrades the accuracy of the attacker's clone model (by up to $40\%$), while minimally impacting the accuracy ($<0.5\%$) for benign users. Compared to existing defenses, our defense has a significantly better security vs accuracy trade-off and incurs minimal computational overhead.
\end{abstract}

\blfootnote{
Preprint. Under submission.\\
}

\ignore{
Flow

1. deep learning has sota accuracy on various tasks
2. dnn used to offer various services, MLaS
3. dnn is valuable asset of the company and is kept confidential
4. in the hands of an adversary the dnn can have a range of consequences
5. Can be used to replicate service and avoid paying for service, impacting business, privacy concerns by leaking confidential data, adversarial attacks vulnerable to safety critical applications
6. There is a need to protect the confidentiality of dnns
7. Unfortunately, several attacks have shown that it is possible to perform model stealing attacks with remarkable accuracy. Moreover it has been demonstrated that it can be done with the following constraints: 1. data limitation 2 black-box access
8.Given the lack of access to data, attacks either make use of synthetic data generated by a small number of 'seed' examples or they use data from a surrogate dataset(such as a dataset of cats for a dog breed classifier). Such attacks make it possible for an adversary to clone 
9. We want to create a defense for such attacks
10. As the adversary is unaware of the training data's distribution,  all existing attacks invariably generate out of distribution (OOD) queries. Fig 2. shows the 
}

\section{Introduction}
The ability of Deep Learning models to solve several challenging classification problems in fields like computer vision and natural language processing has proliferated the use of these models in various products and services such as smart cameras, intelligent voice assistants and self-driving cars. In addition, several companies now employ deep learning models to offer classification as a service to end users who may not have the resources to train their own models.
\begin{figure}[htb]
	\centering
    \centerline{\epsfig{file=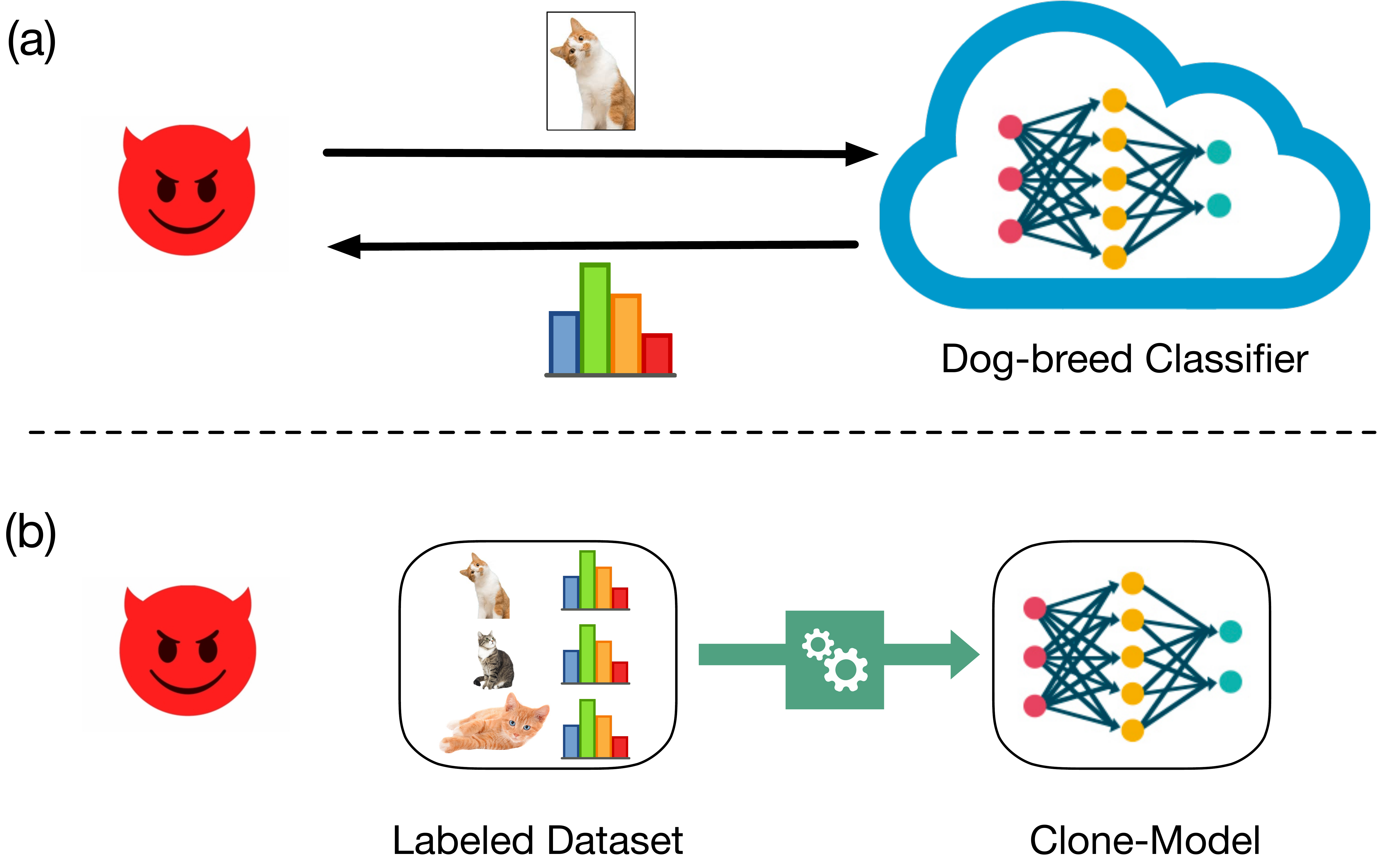, width=0.9\columnwidth}}
	\caption{Model Stealing Attack: (a) An adversary queries the target model (dog-breed classifier) using synthetic/surrogate data (cat images) and constructs a labeled  dataset using the predictions of the model (b) The labeled dataset can then be used to train a clone-model that replicates the functionality of the target model.}
	\vspace{-0.2in}
    \label{fig:knockoff}
\end{figure}
In most of these cases, while the model parameters and the architecture are kept hidden from the end-user, the user is allowed to interact with the model to obtain the classification outputs for the user's inputs. 
The confidentiality of these models is important as the models can be misused in various ways in the hands of an adversary. For instance, an adversary can use the stolen model to offer a competing service which can be detrimental to business. Furthermore, stolen models can be used to craft adversarial examples~\cite{intrigue, exp_harness, practical_bbox, transfer}, creating vulnerability for safety-critical applications such as malware detection and can even leak information about the data used to train the model, causing privacy issues~\cite{model_inversion_confidence, mem_inference}. These issues create a need to protect the confidentiality of machine learning models.

Unfortunately, recent attacks~\cite{knock, practical_bbox} have shown that it is possible for an adversary to carry out model stealing attacks and train a clone model that achieves a classification accuracy that is remarkably close to the accuracy of the target model (up to $0.99\times$). Moreover, these attacks can be performed even when the adversary is constrained in the following ways:

1. The adversary only has \emph{black-box} query access to the model i.e. the attacker can query the model with any input and observe the output probabilities.

2. The adversary is \emph{data-limited} and does not have access to a large number of inputs representative of the training data of the target model. 
\begin{figure}[htb]
	\centering
    \centerline{\epsfig{file=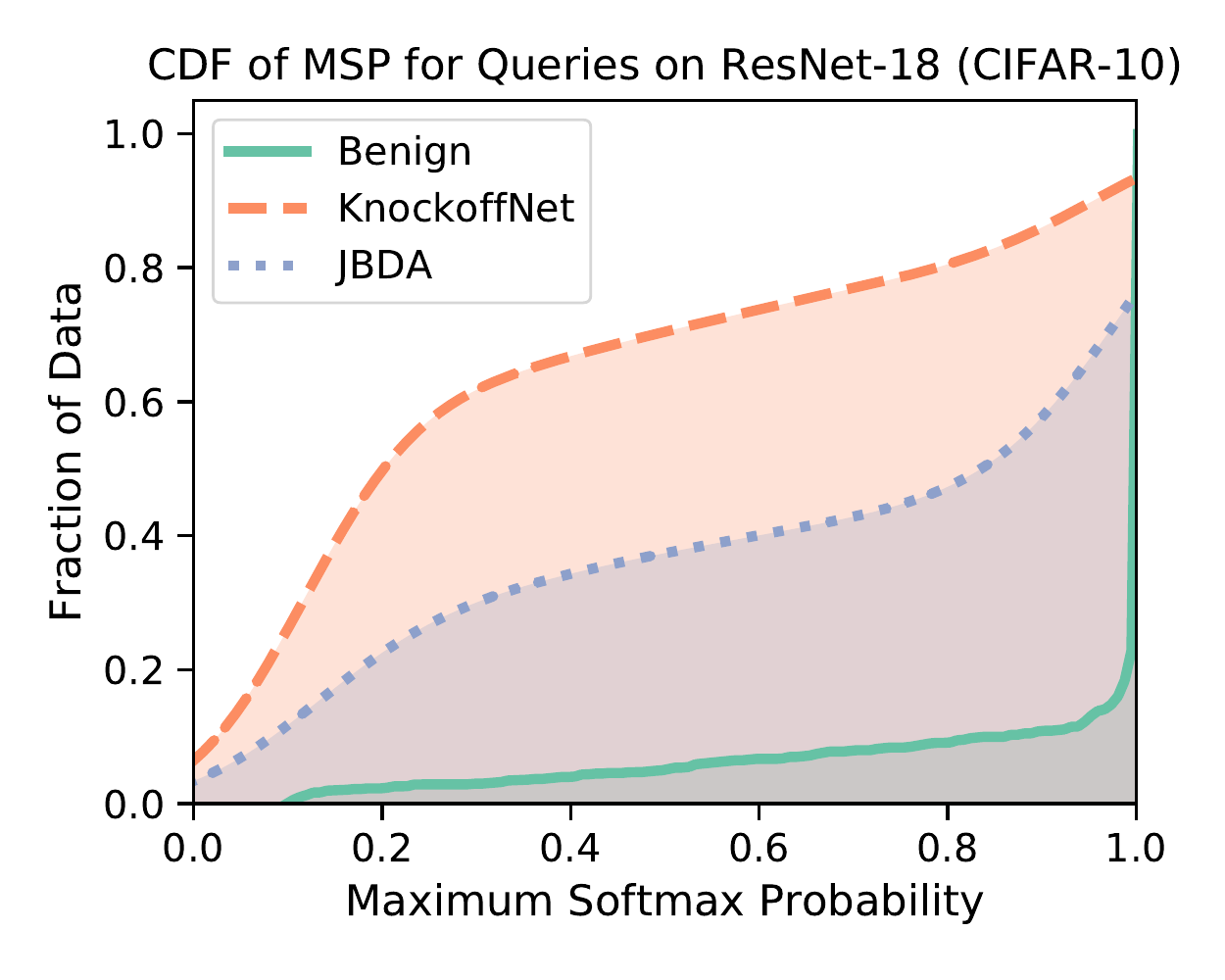, width=0.8\columnwidth}}
	\caption{CDF of Maximum Softmax Probability (MSP) for queries from: (a) Benign User (b) KnockoffNet Attacker (c) Jacobian-Based Dataset Augmentation (JBDA) Attacker. Queries from benign user produce high values of MSP indicating in-distribution data while queries generated from attacks produce low values of MSP indicating out-of-distribution data.}
	\vspace{-0.1in}
    \label{fig:msp_hist}
\end{figure}

Attacks that work under these constraints rely on one of two methods for generating the data necessary for querying the target model: (a) \emph{Synthetic Data}: ~\cite{prada,practical_bbox} produce synthetic data from a small set of in-distribution \emph{seed} examples by iteratively adding heuristic-based perturbations to the seed examples. (b) \emph{Surrogate Data}: Several attacks ~\cite{knock, stealing_ml} simply use a surrogate dataset to query the target model. For example, a cat dataset can be used as the surrogate dataset to query the dog-breed classifier as shown in Fig~\ref{fig:knockoff}a.
A labeled dataset can be constructed from these queries, which can be used by the adversary to train a clone model that mimics the functionality of the target model (Fig~\ref{fig:knockoff}b). 
Such attacks make it viable for an adversary to create a clone of the target model even with limited/no access to the target model's data distribution. The goal of this paper is to propose an effective defense for model stealing attacks carried out by a data-limited adversary with black-box access to the target model.

 We observe that all existing attacks invariably generate Out-Of-Distribution (OOD) queries. One way to check if the data is OOD is by plotting the Maximum Softmax Probability (MSP) of the data produced by the target model. High values of MSP indicate In-Distribution (ID) data and low values indicate OOD data ~\cite{ood}. As an example, we characterize the MSP values using a ResNet-18 network trained on the CIFAR-10 dataset. We plot the CDF of MSP for benign queries sampled from a held-out test set as well as adversarial queries from two representative attacks: 1. Knockoffnets~\cite{knock}, using surrogate data from CIFAR-100 dataset and 2. Jacobian-Based Data Augmentation (JBDA)~\cite{practical_bbox}, which uses synthetic data, in Fig~\ref{fig:msp_hist}. Notice that the CDFs of the queries from both attacks are concentrated towards lower values of MSP, indicating OOD data, compared to the inputs from the benign user which produce high MSP values, implying that the inputs are ID.

Motivated by this observation, we propose \emph{Adaptive Misinformation} (AM) to defend against model stealing attacks. AM selectively sends incorrect predictions for queries that are deemed OOD, while ID queries are serviced with correct predictions. Since a large fraction of the adversary's queries is OOD, this leads to the mislabeling of a significant portion of the adversary's dataset. Training a model on this mislabeled dataset results in a low-quality clone with poor accuracy, reducing the effectiveness of model stealing attacks. Recent works~\cite{deceptive_perturbations, pp} have used a similar insight of misleading the adversary, by injecting perturbations to the predictions of the model.
Compared to these perturbation based defenses, our proposal is more scalable and offers a significantly better trade-off between model accuracy and security due to the following key attributes:



1. \emph{Adaptive Nature:} The adaptive nature of our defense allows using incorrect predictions to selectively service suspicious OOD queries, instead of indiscriminately adding perturbations to the probabilities for all inputs. This results in a better trade-off between model accuracy and security against model stealing attacks.

2. \emph{Reduced Correlation through Misinformation:} Prior works add perturbations to the original prediction in order to mislead the adversary. However, we find that these perturbed predictions remain correlated with the original predictions, leaking information about the original predictions of the model. In contrast, our defense uses an uncorrelated misinformation function to generate incorrect predictions, which reveals no information about the original predictions, resulting in better security.

3. \emph{Low Computational Overhead:} Our proposal only requires a single inference pass with a modest increase in the amount of computation over an undefended model ($<2\times$). In contrast, existing defenses like Prediction Poisoning~\cite{pp} (PP) requires multiple gradient computations and thus incurs several orders of magnitude increase in computational cost and inference latency.

Overall, the contributions of our paper are as follows:

1. We analyze the queries from existing model stealing attacks (KnockoffNets and JBDA) and identify that these attacks produce a large number of OOD queries. We leverage this observation to develop an effective defense.

2. We propose \emph{Adaptive Misinformation} to defend against model stealing attacks. Our defense involves using an out of distribution detector to flag ``suspicious" inputs, potentially from an adversary, and adaptively servicing these queries with incorrect predictions from an auxiliary ``misinformation model" which produces uncorrelated predictions

3. We perform extensive empirical studies to evaluate our defense against multiple model stealing attacks. We plot the security vs accuracy trade-off curve for various datasets and show that, owing to its adaptive nature, our defense achieves a significantly better trade-off compared to prior art. For instance, in the case of the Flowers-17 dataset, AM lowers the clone model accuracy to 14.3\%, compared to the clone accuracy of 63.6\% offered by PP with comparable defender accuracy of 91\%.



\section{Problem Description}

Our problem setting involves a \emph{data-limited adversary} who is trying to perform model stealing attack on a \emph{defender}'s model, just using black-box query access. In this section, we outline the attack and defense objectives. We also provide background on various attacks that have been proposed in literature to perform model stealing for a data-limited adversary.

\subsection{Attack Objective}
The adversary's goal is to replicate the functionality of the defender's model $f(x; \theta)$ by training a clone model $f'(x; \theta')$ that achieves high classification accuracy on the defender's classification task, as shown in Eqn.~\ref{eq:adv_obj}. Here, $P_{def}(X)$ denotes the distribution of data from the defender's problem domain, $\theta$ represents the parameters of the defender's model, and $\theta'$ represents the parameters of the clone model that the attacker is trying to train.

\begin{align} \label{eq:adv_obj}
\max_{\theta'}\mathop{\mathbb{E}}_{x\sim \mathcal{P}_{def}(X)}Acc(f'(x;\theta'))
\end{align}
If the adversary had access to a labeled dataset of inputs sampled from $P_{def}(X)$, the adversary could simply use this to train the clone-model $f'$. However, in a lot of real-world classification problems, the adversary is data-limited and lacks access to a sufficiently large dataset that is representative of $P_{def}(X)$. 

\subsection{Model Stealing Attacks under data limitations}~\label{sec:model_stealing_background}
In the absence of a representative dataset, the adversary can use either \emph{synthetic} or \emph{surrogate} data to query the defender's model. A labeled dataset can be constructed from the predictions obtained through these queries, which can be used to train the clone model $f'$. These methods rely on the principle of \emph{knowledge distillation}~\cite{kd}, where the predictions of a ``teacher" model (defender's model) are used to train a ``student" model (attacker's clone model). We describe ways in which an attacker can generate synthetic and surrogate data to perform model stealing attacks.

\textbf{(a) Synthetic Data:} The adversary starts by training a surrogate model $f'$ using a small \emph{seed} dataset $\mathcal{D}_{seed}$ of ID examples and iteratively augments this dataset with synthetic examples. \emph{Jacobian-Based Data Augmentation} (JBDA) ~\cite{practical_bbox, prada} is one such heuristic for generating synthetic examples. For each input $x\in\mathcal{D}$, this method generates a synthetic example $x'$, by perturbing it using the jacobian of the clone model's loss function: $x' = x + \lambda sign\left(\nabla_x\mathcal{L}\left(f'\left(x;\theta'\right)\right)\right)$. These synthetic examples are labeled using the predictions of the defender's model $y'=f(x')$ and the labeled synthetic examples thus generated: $\mathcal{D}_{syn}=\{x', y'\}$, are used to augment the adversary's dataset: $\mathcal{D}_{seed}=\mathcal{D}_{seed}\cup\mathcal{D}_{syn}$ and retrain $f'$. 

\textbf{(b) Surrogate Data:} Recent works ~\cite{knock, stealing_ml} have shown that it is possible to use a surrogate distribution $P_{sur}(X')$, which is dissimilar from $P_{def}(X)$, to steal the functionality of black-box models. The adversary can use inputs $x' \sim P_{sur}(X')$ to query the defender's model and obtain the prediction probabilities $y'=f(x')$. The labeled data thus obtained: $\mathcal{D}_{sur}=\{x', y'\}$, can be used as the surrogate dataset to train the clone-model $f'$.

Such methods enable model stealing attacks, despite the data limitations of the adversary, posing a threat to the confidentiality of the defender's black-box model.

\subsection{Defense Objective}

The defender's aim is to prevent an adversary from being able to replicate the functionality of the model. Thus the defender's objective involves minimizing the accuracy of the cloned model $f'$ trained by the adversary (Eqn.~\ref{eq:def_obj}).
\begin{align} \label{eq:def_obj}
&\min\mathop{\mathbb{E}}_{x\sim \mathcal{P}_{def}(X)}[Acc(f'(x;\theta'))]
\end{align}
The defender is also constrained to provide high classification accuracy to benign users of the service in order to retain the utility of the model for the classification task at hand. We formalize this by stating that the classification accuracy of the model for in-distribution examples has to be above a threshold $T$.
\begin{align} \label{eq:def_constraint}
\mathop{\mathbb{E}}_{x\sim \mathcal{P}_{def}(X)}[Acc(f(x;\theta))] \geq T
\end{align}
Eqn.~\ref{eq:def_obj},~\ref{eq:def_constraint} describe a constrained optimization problem for the defender. This formulation of the problem allows the defense to trade off the accuracy of the model for improved security, as long as the accuracy constraint (Eqn.~\ref{eq:def_constraint}) is satisfied. We term defenses that work within these constraints as \emph{accuracy-constrained} defenses. Our proposed defense falls under this framework and allows improvement in security at the cost of a reduction in classification accuracy.

\ignore{
Flow

Stateful Detection
These methods aim to detect adversarial attacks by analyzing the distribution of queries from individual users. [x] for instance uses distance between successive queries to detect adversarial attacks based on the assumption that adversarial users send highly correlated queries. Unfortunately, such methods require the defender to maintain a history of past queries, limiting scalability. Moreover, these defenses are ineffective against attacks involving multiple colluding adversarial users and attacks that use surrogate datasets which do not use correlated queries.

Perturbation based Defense
1. Accuracy preserving perturbations
2. Accuracy-Constrained Perturbations

}
\section{Related Work}

We discuss the various defenses against model stealing attacks that have been proposed in literature. Existing defenses can broadly be categorized into \emph{Stateful Detection Defenses} and \emph{Perturbation Based Defenses}. 

 \subsection{Stateful Detection Defenses}
 
 Several works ~\cite{prada, stateful} have proposed analyzing the distribution of queries from individual users to classify the user as adversarial or benign. For instance, ~\cite{prada} uses the $L_2$ distance between successive queries to detect adversarial attacks based on the assumption that adversarial users send highly correlated queries. Unfortunately, such methods requires the defender to maintain a history of past queries limiting scalability. Moreover, these defenses are ineffective against adaptive attacks, attacks involving multiple colluding adversarial users and attacks that use surrogate datasets, which do not have correlated queries.

\subsection{Perturbation-Based Defenses}
In an undefended setting, the attacker has reliable access to the predictions of the target model $f$ for any arbitrary input $x$. Perturbation-based defenses modify the original prediction of the model $y=f(x;\theta)$ to produce a perturbed prediction $y'$, preventing the adversary from having reliable access to the target model's predictions.  
Consequently, training a clone model with the surrogate dataset: $\{x,y'\}$ formed by the adversary results in a low-quality clone model with reduced accuracy. 
There are several recent defenses that work under different constraints for generating perturbed predictions. They can be broadly categorized into defenses that preserve the accuracy of the model and defenses that trade-off accuracy for security. We briefly describe each of these works before detailing our solution in the following section.

\subsubsection{Accuracy Preserving Defenses} 
These defenses ensure that the accuracy of the model on ID examples is unchanged after adding the perturbations. For instance, ~\cite{deceptive_perturbations} constrains the perturbation to leave the top-1 class of the perturbed output unchanged i.e $argmax(y'_i) = argmax(y_i)$.
This preserves the prediction accuracy of the model, while removing the information present in the probabilities outputted by the defender's model. Another defense ~\cite{stealing_ml}, which works using the same principle, avoids exposing the output probabilities by only sending the hard labels of the top-1 or top-K classes while servicing requests. Both these defenses prevent the adversary from accessing the true prediction probabilities either implicitly or explicitly, while retaining the accuracy of the defender's model. Unfortunately, subsequent works have shown that the effectiveness of these defenses are limited as the adversary can still use the top-1 prediction of the model to perform model stealing attacks~\cite{pp}.
 \subsubsection{Accuracy-Constrained Defenses} 
Unlike accuracy preserving attacks, accuracy-constrained defenses do not require the classification accuracy to be retained after perturbing the predictions of the model. This allows the defender to trade off model accuracy for better security by injecting a larger amount of perturbation to the predictions of the model. However, the amount of perturbation that can be injected is bound by the accuracy constraint (Eqn.~\ref{eq:def_constraint}), which ensures that the accuracy of the model is above a specified threshold for ID examples. Prediction Poisoning~\cite{pp} (PP) is a recent work that proposes such an accuracy-constrained defense, whereby the defender perturbs the prediction of the model as shown in Eqn.~\ref{eq:predpoison}
 \begin{align} \label{eq:predpoison}
y' = (1-\alpha) f(x;\theta) + \alpha \eta
\end{align}
The perturbed output $y'$ is computed by taking a weighted average between the predictions of the true model $f$ and a poisoning probability distribution $\eta$. The poisoning distribution $\eta$ is computed with the objective of mis-training the adversary's clone model. This is done by maximizing the angular deviation between the weight gradients of the perturbed prediction with that of the original predictions of the model. $\alpha$ is  a tunable parameter that controls the weightage given to the poisoned distribution and the true output of the model. Thus, increasing $\alpha$ allows the defender to trade off the accuracy of the model for increased security against model stealing attacks by increasing the amount of perturbation injected into the model's original predictions.
\section{Our Proposal: Adaptive Misinformation}
Our paper proposes \emph{Adaptive Misinformation} (AM), an accuracy-constrained defense to protect against model stealing attacks. Our defense is based on the observation that existing model stealing attacks generate a large number of OOD examples to query the defender's model. This is because the adversary is data-limited and does not have access to a large dataset representative of the defender's training dataset. Our defense takes advantage of this observation by adaptively servicing queries that are OOD with \emph{misinformation}, resulting in most of the attacker's queries being serviced with incorrect predictions.
 
Consequently, a large fraction of the attacker's dataset is mislabeled, degrading the classification accuracy of the clone model trained on this poisoned dataset. 
Compared to existing perturbation based defenses like PP, our proposal has the following distinguishing qualities: 

1. AM selectively modifies the predictions only for OOD queries, leaving the predictions unchanged for ID inputs. This is in contrast to prior works, which perturb the predictions indiscriminately for all inputs 

2. In existing perturbation based defenses, there is a significant amount of correlation between and perturbed prediction $y'$ and the original prediction $y$, which leaks information about the original prediction, that can be exploited by an adversary (discussed further in Section~\ref{sec:knockoff_exp}). In contrast, AM ensures that $y'$ is uncorrelated with $y$ for OOD queries and therefore avoids leaking information about the original predictions.

3. PP requires expensive gradient computations to determine the perturbations. In contrast, AM has low computational overheads and just requires evaluation of an auxiliary \emph{misinformation} model.

\begin{figure}[htb]
	\centering
    \centerline{\epsfig{file=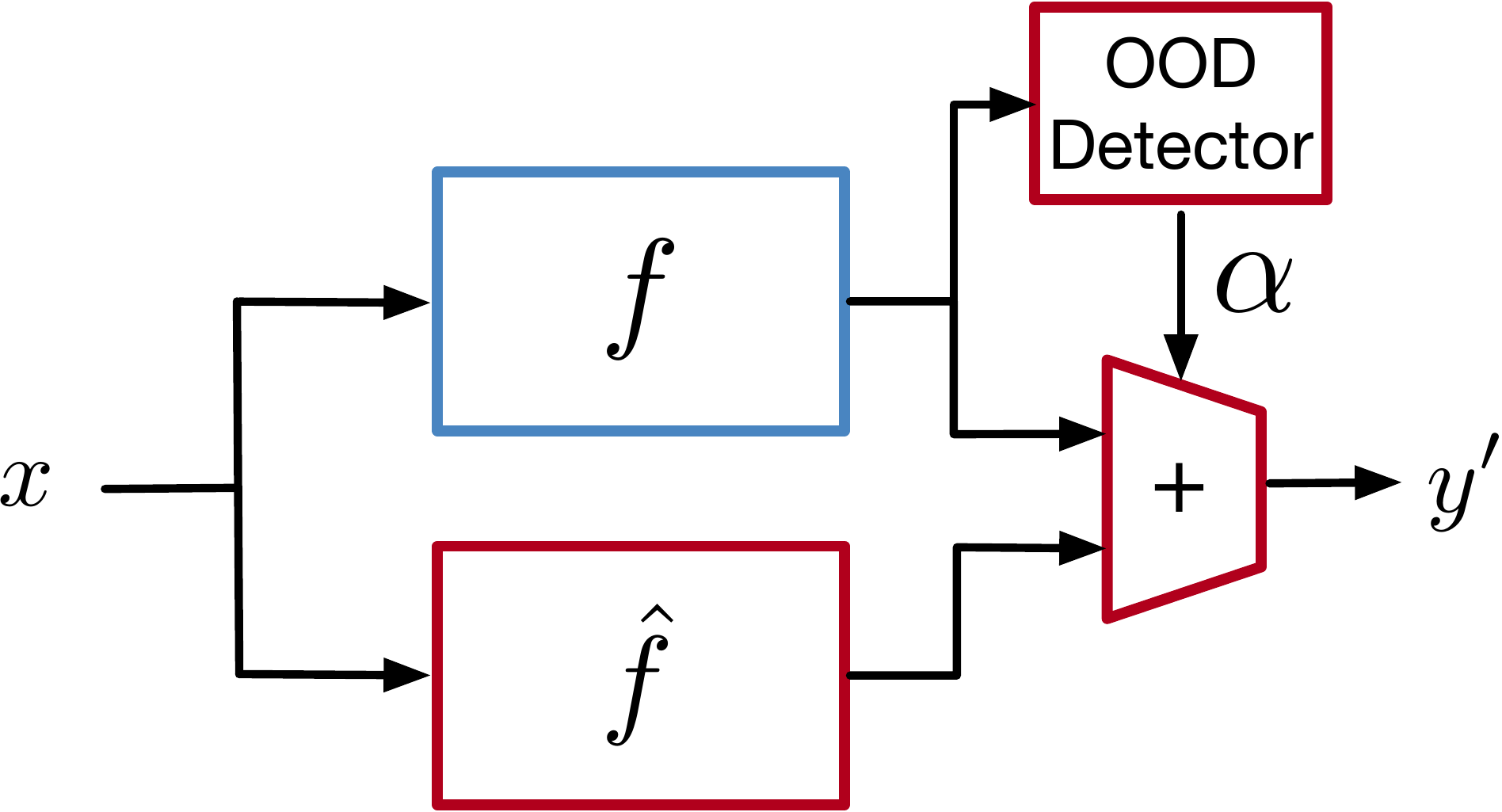, width=0.8\columnwidth}}
    \vspace{0.1 in}
	\caption{Adaptive Misinformation: We use an OOD detection mechanism to selectively service OOD inputs with the predictions of the misinformation function $f'$, while the ID inputs are serviced with the original predictions of the model $f$.}
    \label{fig:adaptive_misinformation}
    \vspace{-0.1in}
\end{figure}

These advantages allow our defense to achieve a better trade-off between classification accuracy and security compared to existing defenses, with a low computational overhead. Fig.~\ref{fig:adaptive_misinformation} shows the block diagram of our proposed Adaptive Misinformation defense. In addition to the defender's model $f$, there are three components that make up our defense: (1) An OOD detector (2) A misinformation function ($f'$) (3) A mechanism to gradually switch between the predictions of $f$ and $f'$ depending on the input.

For an input query $x$, AM first determines if the input is ID or OOD. If the input is ID, the user is assumed to be benign and AM uses the predictions of $f$ to service the request. On the other hand, if $x$ an OOD input, the user is considered to be malicious and the query is serviced using the incorrect predictions generated from $\hat{f}$. In the remainder of this section, we explain the different components of our defense in more detail.  
\subsection{Out of Distribution Detector} ~\label{sec:ood}
Out of Distribution detection is a well-studied problem in deep learning~\cite{ood, ood1, ood2, ood3, ood4, ood5}, where the objective is to determine if an input received by the model during testing is dissimilar to the inputs seen during training. This can be used to detect and flag anomalous or hard to classify inputs which might require further examination or human intervention. A simple proposal to detect OOD examples~\cite{ood} involves using the Maximum Softmax Probability (MSP) of the model. For a model that outputs a set of $K$ output probabilities $\{y_i\}_{i=1}^K$ for an input $x$, OOD detection can be done by thresholding the MSP as shown in Eqn.~\ref{eq:detection}.
\begin{align} \label{eq:detection}
Det(x) = 
\begin{cases}
    ID & \text{if } \max_i(y_i) > \tau\\
    OOD & otherwise
\end{cases}
\end{align}
The idea here is that the model produces confident predictions on ID inputs, similar to the ones seen during training and less confident predictions on OOD examples that are dissimilar to the training dataset.
Outlier Exposure~\cite{oe} is a recent work that improves the performance of the threshold-based detector by exposing the classifier to an auxilary dataset of outliers $\mathcal{D}_{out}$. The model is trained to produce uniform probability distribution ($\mathcal{U}$) on inputs from $\mathcal{D}_{out}$ by adding an extra term to the loss function during training as shown in Eqn.~\ref{eq:oe}.
\begin{align} \label{eq:oe}
\mathbb{E}_{(x, y) \sim \mathcal{D}_{\text {in }}}\left[\mathcal{L}\left(f\left(x\right), y\right)\right]+\lambda \mathbb{E}_{x^{\prime} \sim \mathcal{D}_{\text {out}}}\left[\mathcal{L}\left(f\left(x^{\prime}\right), \mathcal{U}\right)\right]
\end{align}
This ensures that the model produces accurate and confident predictions for inputs sampled from $\mathcal{D}_{in}$, while OOD examples produce less confident predictions, improving the ability of the detector to distinguish them. We train the defender's model with outlier exposure and use a threshold-based detector in our defense to perform OOD detection.
\subsection{Misinformation Function}
For queries which are deemed OOD by the detector, we want to provide incorrect predictions that are dissimilar to the predictions of the true model in order to deceive the adversary. We obtain the incorrect predictions by using a \emph{misinformation function} $\hat{f}$. We train $\hat{f}$ by minimizing the \emph{reverse cross entropy loss} as shown in Eqn.~\ref{eq:poison}.
\begin{align} \label{eq:poison}
\min_{\hat{\theta}}\mathop{\mathbb{E}}_{(x,y)\sim \mathcal{D}_{\text{in}}}[ \mathcal{L}\big((1-\hat{f}(x;\hat{\theta})),y\big)]
\end{align}
Minimizing the reverse cross entropy loss trains $\hat{f}$ to produce incorrect predictions. We use this model to provide misleading information to OOD queries, making it harder for an adversary to train a clone model that obtains high accuracy on the classification task.

\subsection{Adaptively Injecting Misinformation}~\label{sec:adaptively_injecting_perturbations}
Finally, we need a mechanism to gradually switch between the outputs of the defender's model ($f$) and the misinformation model ($\hat{f}$), depending on whether the input $x$ is ID or OOD. In order to achieve this, we first pass $x$ through an OOD detector, which simply requires computing the maximum softmax probability $y_{max}$ of all the output classes produced by $f$.
\begin{align} \label{eq:ood}
y_{max} = \max_i({y_i})
\end{align}
A larger value of $y_{max}$ indicates that the input is ID, while a smaller value indicates an OOD input. We use a threshold $\tau$ to classify between ID and OOD inputs as shown in Eqn.~\ref{eq:detection}. The predictions of $f$ and $\hat{f}$ are combined by using a reverse sigmoid function $S(x)$ to produce the final output probabilities $y'$ as shown in Eqn.~\ref{eq:combination}.
\begin{align}
y' = (1-\alpha&) f(x;\theta) + \left(\alpha\right)\hat{f}(x;\hat{\theta}) \label{eq:combination}\\
where\ \ \ \ \  \alpha &= S(y_{max}-\tau) \label{eq:ood_threshold}\\
S(z) &= \frac{1}{1+e^{\nu z}}  \label{eq:sigmoid}
\end{align}
Thus for an ID input, with $y_{max}>\tau$, we obtain $\alpha < 0.5$ with $y' \to f(x;\theta)$ as $\alpha \to 0$. Similarly, for an OOD input with $y_{max}<\tau$, we obtain $\alpha>0.5$ with $y' \to \hat{f}(x;\hat{\theta})$ as $\alpha \to 1$. $\nu$ in Eqn.~\ref{eq:sigmoid} indicates the growth rate of the sigmoid. We set $\nu=1000$ for all of our experiments. Thus, an adversary accessing the model with OOD inputs obtains the predictions of $f'$ instead of the true predictions of model $f$, while inputs from benign users of the service sending ID queries would be serviced by $f$. This results in the adversary's dataset containing examples which have been mislabeled, leading to a degradation in the accuracy of the clone model trained on this data.  

\textbf{Security vs Accuracy Trade-off:} The OOD detector has a trade-off between true and false positive rates. In general, by lowering the value of the detector threshold $\tau$, we can increase the number of OOD inputs classified correctly (true positive rate), which improves security as more OOD queries are serviced with misinformation. However, this also results in a higher number of ID inputs misclassified as OOD (false positive rate), leading to a greater number of ID inputs being serviced with misinformation, degrading the accuracy of the defender's model for benign ID examples. By appropriately setting the value of $\tau$, the defender can pick a trade-off point between security and accuracy that satisfies the accuracy-constraint (Eqn.~\ref{eq:def_constraint}).

\section{Experiments}
We perform experiments to evaluate our defense against various model stealing attacks.  Additionally, we compare AP against existing defenses and show that our defense offers better protection against model stealing compared to prior art. We describe our experimental setup followed by the results in this section. 
\subsection{Setup}
Our experimental setup involves a defender who hosts a model $f$, trained for a specific classification task. The attacker aims to produce the clone model $f'$, which achieves high classification accuracy on the same classification task. We briefly describe the classification tasks as well as the attacks and defenses that we use in our evaluations.

\textbf{Datasets and model architecture: } We focus on vision based classification tasks using DNNs in our experiments. Table~\ref{table:datasets} lists the various datasets and model architectures used to train the defender's model $f$ as well as the test accuracy achieved by the models on the corresponding classification task. As mentioned in section~\ref{sec:ood}, we train our model with \emph{outlier exposure}~\cite{oe} to improve the performance of OOD detection. For this purpose, we use KMNIST~\cite{kmnist} for MNIST and FashionMNIST, ImageNet1k for CIFAR-10, and Indoor67~\cite{indoor67} for Flowers-17~\cite{flowers} as the outlier datasets.
\begin{table}[htb]
\begin{center}
\begin{tabular}{|l|l|c|}
\hline
Dataset           & DNN Architecture     & Accuracy(\%) \\
\hline\hline
MNIST             & LeNet     & 99.4        \\
FashionMNIST      & LeNet     & 91.47        \\
CIFAR-10          & ResNet-18 & 93.6        \\
Flowers-17 & ResNet-18 & 98.2       \\
\hline
\end{tabular}
\end{center}
\caption{Datasets and model architectures used to train the defender's model}
\label{table:datasets}
\vspace{-0.2in}
\end{table}

\textbf{Attacks: }
We evaluate our defense against two representative model stealing attacks:

\emph{1. KnockoffNets~\cite{knock}:} This attack uses surrogate data to perform model stealing as described in section~\ref{sec:model_stealing_background}. We use EMNISTLetters/EMNIST/CIFAR-100/ImageNet1k as the surrogate datasets to attack the MNIST/FashionMNIST/CIFAR-10/Flowers-17 models respectively. We assume a query budget of 50000 examples and train all clone models for 50 epochs. 

\emph{2. Jacobian-Based Dataset Augmentation (JBDA)~\cite{prada,practical_bbox}:} This attack constructs a synthetic dataset by iteratively augmenting an initial set of \emph{seed} examples with perturbed examples constructed using the jacobian of the clone model's loss function. We use a seed dataset of 150 examples with 6 rounds of augmentation to construct the adversary's dataset. Between each augmentation round, the clone model is trained for 10 epochs and $\lambda$ is set to 0.1.

To improve the efficacy of the attacks, we use the same model architecture as the defender's model to train the clone model. Additionally, for the ResNet-18 models, we initialize the weights using a network pre-trained on the ImageNet dataset. We use a learning rate of 0.1 and 0.001 for LeNet and ResNet-18 models respectively. 
\begin{figure*}[htb]
	\centering
    \centerline{\epsfig{file=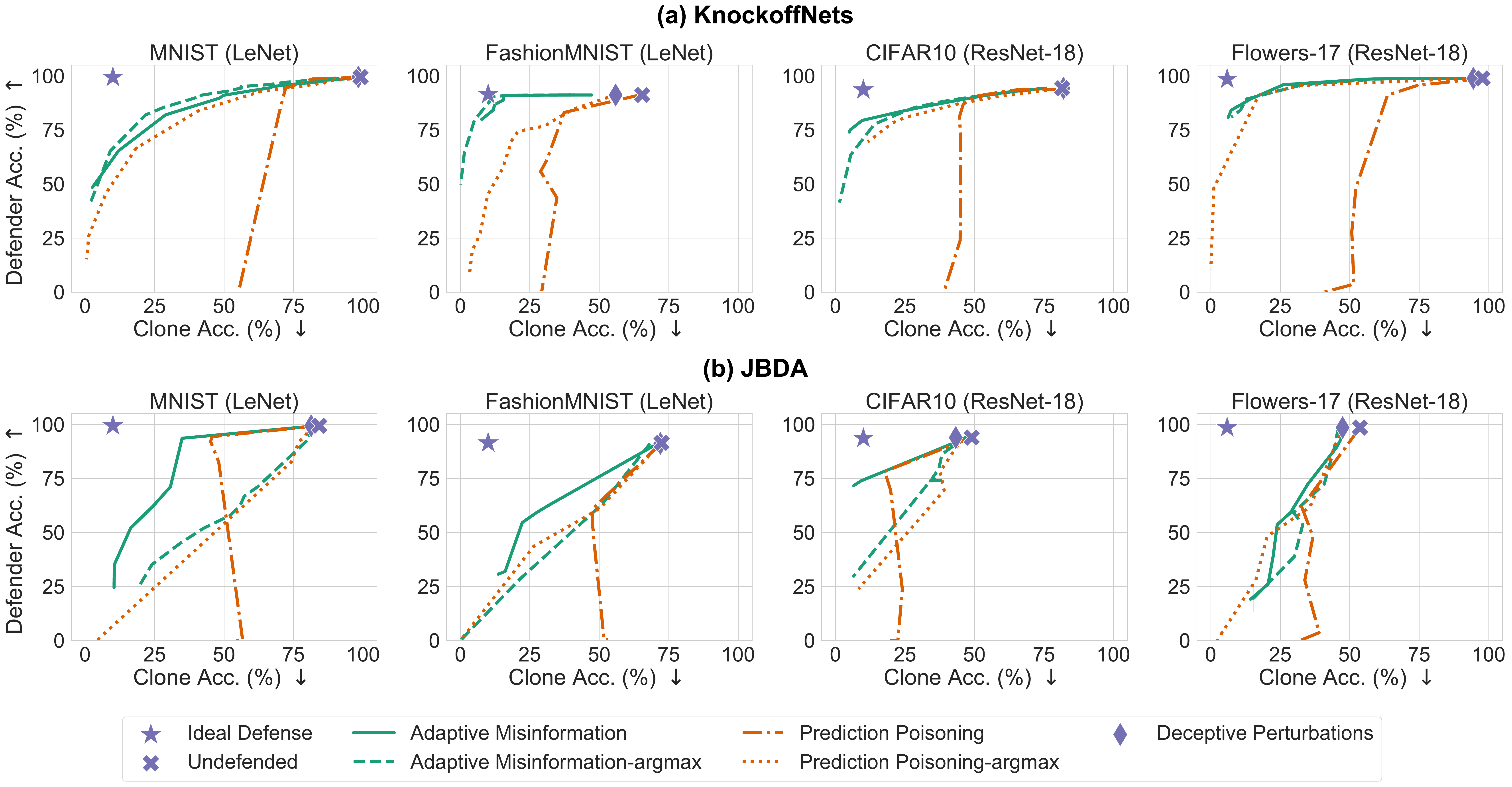, width=\textwidth}}
	\caption{ Defender Accuracy vs Clone Accuracy trade-off for defenses evaluated against two attacks: (a) \emph{KnockoffNets} (b) \emph{Jacobian Based Dataset Augmentation}. Perturbation based defenses can improve security (lower clone model accuracy) at the expense of reduced defender accuracy. Our proposal \emph{Adaptive Misinformation} offers a better trade-off compared to existing defenses. E.g. in case of Flowers-17 dataset with KnockoffNets attack, PP achieves a clone accuracy of 63.6\% with a defender accuracy of 91.1\%. In comparison, AM yields a much lower clone accuracy of 14.3\% (-49.3\%) for the same defender accuracy, significantly improving the trade-off compared to PP.}
    \label{fig:results}
    \vspace{-0.15in}
\end{figure*}

\textbf{Comparison with existing Defenses:}
Only a small number of defenses currently exist for model stealing attacks. We compare our defense against two recent perturbation-based defenses:

\emph{1. Deceptive Perturbation (DP):} This is a accuracy-preserving defense that adds deceptive perturbations to the predictions of the model but leaves the top-1 class unchanged\cite{deceptive_perturbations}. Injecting perturbations removes information about prediction probabilities but preserves information about the \emph{argmax} prediction of the model.

\emph{2. Prediction Poisoning (PP):} This is an accuracy-constrained defense that perturbs the predictions of the model with an objective of mistraining the adversary's clone model\cite{pp}. Increasing the amount of perturbation allows the defender to trade off the model accuracy for increased security, similar to our defense.
\subsection{Results}
Our defense allows the defender to trade off the defender's model accuracy for increased security against model stealing attacks by varying the threshold $\tau$ of the OOD detector as described in section~\ref{sec:adaptively_injecting_perturbations}. We measure security by the accuracy of the clone model trained using model stealing attacks, with a lower clone-model accuracy indicating better security. We plot this trade-off curve of defender's model accuracy vs clone model accuracy evaluated against different attacks and show that our defense offers a better trade-off compared to existing defenses for various classification tasks.
\subsubsection{KnockoffNets Attack}~\label{sec:knockoff_exp}
Figure~\ref{fig:results}a shows the trade-off curve of Adaptive Misinformation (AM) evaluated against the \emph{KnockoffNets} attack. Our results show that AM is able to reduce clone model accuracy significantly, with only a small degradation in defender model accuracy. Additionally, we compare our results with the trade-offs offered by two existing defenses: Prediction Poisoning (PP) and Deceptive Perturbations (DP). Note that PP allows a variable amount of perturbation to be added, leading to a trade-off curve whereas DP has a fixed perturbation leading to a single trade-off point. We also plot the trade-off points for an ideal defense and an undefended model for reference. 

\textbf{Comparison with PP:} Our results show that for a given defender accuracy, AM has a lower clone accuracy compared to PP for all datasets, offering a better trade-off between security and accuracy. For instance, AM lowers clone accuracy by 49.3\% compared to PP with comparable defender accuracy (91.1\%) in the case of the Flowers-17 dataset. We highlight and explain two key differences between the trade-off curves of AM and PP:

1. For PP, as we increase security (reduce clone model accuracy), the accuracy of the defender's model declines sharply to $0\%$, while AM retains high defender model accuracy. This is because PP indiscriminately perturbs predictions for all queries. As the amount of perturbation is increased, the top-1 class of $y'$ changes to an incorrect class, leading to a steep drop in the classification accuracy of benign inputs. Our defense avoids this problem by using an adaptive mechanism that only modifies the probabilities for OOD queries, allowing ID queries to retain a high classification accuracy.

\begin{figure}[thb]
	\centering
    \centerline{\epsfig{file=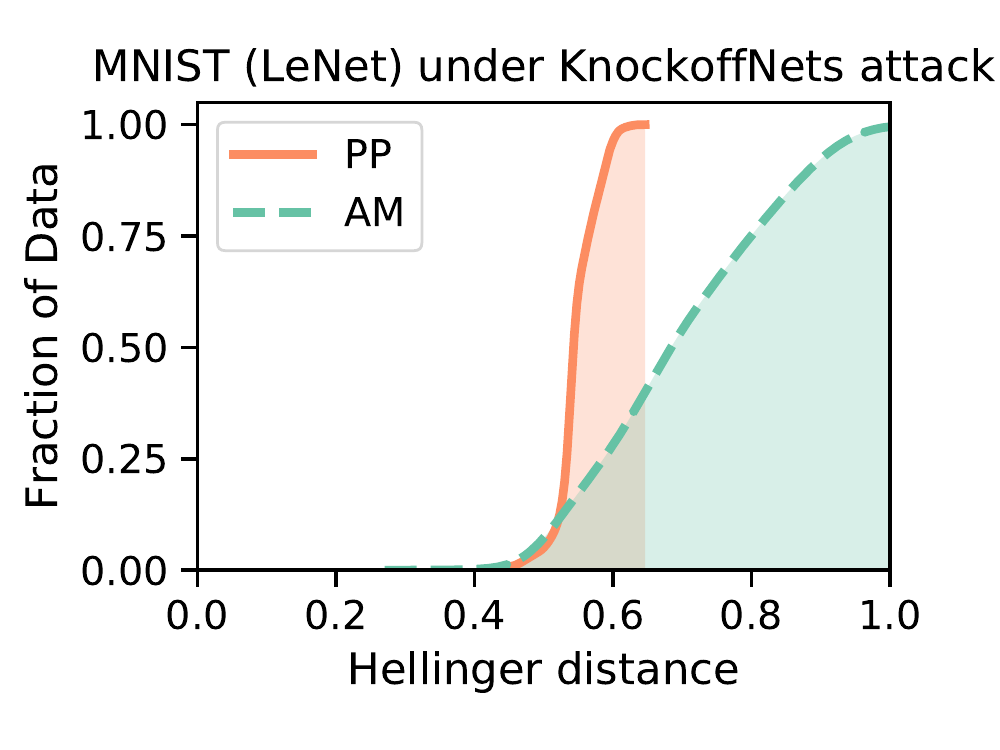, width=0.8\columnwidth}}
    \vspace{-0.1in}
	\caption{CDF of Hellinger distance between the true prediction $y$ and poisoned predictions $y'$, computed for AP and PP for a LeNet model trained on MNIST under KnockoffNets attack, with comparable defender accuracy. The Hellinger distance is larger for AP compared to PP indicating less correlation between $y'$ and $y$}
    \label{fig:hell_dist}
    \vspace{-0.2in}
\end{figure}

2. For PP, even as the defender accuracy falls to $0\%$, the clone accuracy continues to be high (close to $50\%$ for MNIST and Flowers-17). This is because there is a high correlation between the original predictions $y$ and the perturbed predictions $y'$. We can quantify the correlations between $y$ and $y'$ by using Hellinger distance. We plot the CDF of Hellinger distance for a LeNet model (trained with MNIST) under KnockoffNets attack in Fig~\ref{fig:hell_dist}, comparing AP and PP for the same defender accuracy (92\%). 
We find that the predictions of PP have lower Hellinger distance indicating a higher correlation with the true predictions of the defender's model. Sending correlated predictions allows the adversary to learn a clone model with higher accuracy. In contrast, our defense avoids leaking information about the predictions of the original model $f$ by switching to the predictions of the misinformation model $\hat{f}$ when an OOD input is encountered. Therefore, by using uncorrelated probabilities to service OOD queries, AP can offer higher security without severely degrading the defender's model accuracy.

\textbf{Comparison with DP:} For the DP defense, since the top-1 class of the model remains unchanged after adding perturbations, the optimal strategy for the adversary is to use argmax labels from the perturbed predictions to train the clone model. Our results show that DP only marginally improves security compared to an undefended model. In contrast, our defense is able to lower the clone model accuracy significantly. We also evaluate AP and PP with the attacker using an argmax-label strategy. We find this strategy to be less effective for the attacker, as it results in a lower accuracy compared to using the model's predictions to train the clone-model.

\subsubsection{Jacobian Based Dataset Augmentation Attack}~\label{sec:jbda_exp}
Figure~\ref{fig:results}b shows the trade-off curve for the JBDA attack. We find that this attack produces clones with lower accuracies compared to the KnockoffNets attack. The results for the PP defense shows that the defender accuracy quickly drops to 0\%, even as the clone accuracy remains high, similar to  the KnockoffNets attack. Our defense does not suffer from this problem and offers a better trade-off compared to PP. Additionally, we find that using the argmax labels offers a better clone model accuracy for this attack depending on the trade-off point. In this case, AM has a comparable or slightly better trade-off curve compared to PP. As before, the security offered by the DP defense is marginally better than the undefended case, provided the attacker uses argmax labels to train the clone model.

\section{Discussions on Adaptive Attacks}

Since security is a two-player game, it is important to identify and address weaknesses in any defense. In this section, we discuss adaptive attacks against our defense and provide simple solutions that can prevent such attacks.

\textbf{Can the defense mechanism also be treated as part of the black box to perform model stealing attack? :} Given infinitely many examples, an adversary would be able to clone the entire model, including the defense. However, in order to train a high accuracy clone model with a limited query budget, the adversary needs to maximize the number of inputs that get serviced by $f$. Since our defense returns the predictions of $f$ only for in-distribution inputs, just a small fraction of the adversary's queries (which are misclassified as in-distribution by the OOD detector) get serviced by $f$. In the absence of a way to reliably generate in-distribution inputs, the adversary would require a much larger query budget compared to other defenses to reach the desired level of clone accuracy. Furthermore, this would expose the adversary to other detection mechanisms. E.g. a user sending a large fraction of OOD examples can be blacklisted as an adversarial user.

\textbf{Can the adversary distinguish when the inputs are being serviced by $f$ vs $\hat{f}$? :} One way for an adversary to improve the quality of the clone model is to identify and train the clone model only on queries that have been serviced by $f$ i.e. $\alpha \approx 0$. For an input to be serviced by $f$, it has to be classified as an ID example producing high MSP on $f$. Thus, the adversary can potentially use the confidence of the top-1 class as an indication of when the input is serviced by $f$. While this can improve the accuracy of the clone model, the adversary would still need a much larger query budget since only a small fraction of the adversary's queries are serviced by $f$. Additionally, we can easily prevent such detection by smoothing the posterior probabilities of $f$ or sharpening the probabilities of $\hat{f}$ to make the distribution of MSP identical between the outputs of $f$ and $\hat{f}$.
\section{Conclusion}
We propose \emph{Adaptive Misinformation} to defend against black-box model stealing attacks in the data-limited setting. We identify that existing model stealing attacks invariably use out of distribution data to query the target model. Our defense exploits this observation by identifying out of distribution inputs and selectively servicing such inputs with incorrect predictions (misinformation). Our evaluations against existing attacks show that AM degrades clone model accuracy by up to $40\%$ with a minimal impact on the defender accuracy ($<0.5\%$). In addition, our defense has lower computational overhead ($<2\times$) and significantly better security vs accuracy trade-off (up to 49.3\% reduction in clone accuracy) compared to existing defenses.
\section*{Acknowledgements}
We thank our colleagues from the Memory Systems Lab for their feedback. We also gratefully acknowledge the support of NVIDIA Corporation with the donation of the Titan V GPU used for this research.

{\small
\bibliographystyle{ieee_fullname}
\bibliography{egbib}
}

\end{document}